\pdfoutput=1

\documentclass[11pt]{article}

\usepackage[final]{acl}

\usepackage{times}
\usepackage{latexsym}
\usepackage{xurl}
\usepackage{comment}

\usepackage[T1]{fontenc}

\usepackage[utf8]{inputenc}

\usepackage{microtype}

\usepackage{inconsolata}
\usepackage{booktabs}
\usepackage{graphicx}
\usepackage{multirow}
\usepackage{tipa}
\usepackage{xspace}

\newcommand{\repo}{\url{https://github.com/g8a9/building-bridges-gender-fair-german-mt}\xspace}

\newcommand{\graz}{$^{\clubsuit}$}
\newcommand{\ist}{$^{\spadesuit}$}
\newcommand{\hamburg}{$^{\diamondsuit}$}

\title{Gender-fair Machine Translation: A Benchmark}
\title{Investigating Gender-Fair Machine Translation to German}
\title{\textsc{Ge}$^2$\textsc{MT}: An English to German Dataset for Evaluating Gender-Fair Machine Translation}
\title{Building Bridges: A Dataset for Evaluating Gender-Fair Machine Translation into German}

\author{Manuel Lardelli\graz$^*$, Giuseppe Attanasio\ist$^*$, Anne Lauscher\hamburg \\ \\
  \graz~University of Graz,  Austria \\
  \ist~Instituto de Telecomunicações, Lisbon, Portugal \\ 
  \hamburg~University of Hamburg, Germany \\
  \texttt{\href{mailto:manuel.lardelli01@gmail.com}{manuel.lardelli01@gmail.com}}
}

\begin{document}
\maketitle
\begin{abstract}
The translation of gender-neutral person-referring terms (e.g., \emph{the students}) is often non-trivial.
Translating from English into German poses an interesting case---in German, person-referring nouns are usually gender-specific, and if the gender of the referent(s) is unknown or diverse, the generic masculine (\emph{die Studenten (m.)}) is commonly used. 
This solution, however, reduces the visibility of other genders, such as women and non-binary people. To counteract gender discrimination, a societal movement towards using gender-fair language exists (e.g., by adopting neosystems). However, gender-fair German is currently barely supported in machine translation (MT), requiring post-editing or 
manual translations. We address this research gap by studying gender-fair language in English-to-German MT. Concretely, we enrich a community-created gender-fair language dictionary and sample multi-sentence test instances from encyclopedic text and parliamentary speeches.
Using these novel resources, we conduct the first benchmark study involving two commercial systems and six neural MT models for translating words in isolation and natural contexts across two domains. 
Our findings show that most systems produce mainly masculine forms and rarely gender-neutral variants, highlighting the need for future research. 
We release code and data at \repo.

\def\thefootnote{*}\footnotetext{Equal contribution.}\def\thefootnote{\arabic{footnote}}


\end{abstract}

\section{Introduction}
Gender equality is one of the United Nation's sustainable development goals.\footnote{\url{https://sdgs.un.org/goals/goal5}}
As psychological research shows that linguistic forms influence the mental representation of 
gender identities~\citep{sczesny2016can}, many organizations are officially adopting \emph{gender-fair language (GFL)}.\footnote{See, for instance, this recommendation by the European Parliament: \url{http://www.europarl.europa.eu/RegData/publications/2009/0001/P6_PUB(2009)0001_EN.pdf}}

\begin{figure}[!t]
    \centering
    \includegraphics[width=0.95\linewidth]{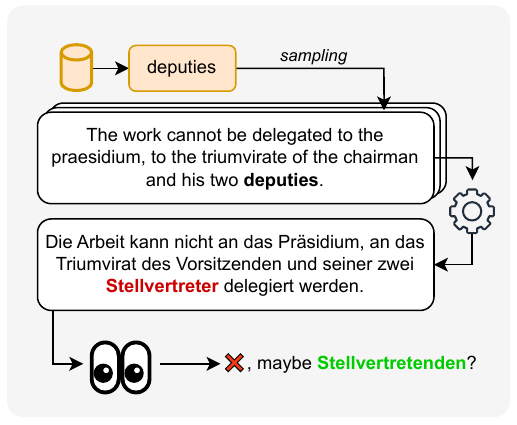}
    \caption{\textbf{Study overview.} We collect English person nouns (yellow, top box) and sample passages representing their mentions in context. We translate those passages with MT systems (white, central boxes) and conduct a human as well as an automatic evaluation on gender forms (bottom) used in German translations.}
    \label{fig:enter-label}
\end{figure}

Towards reaching 
equality and 
inclusion, language technology should account for GFL. 
In this context, recent research in natural language processing (NLP) explores issues around machine translation~\citep[MT; e.g.,][]{piergentili-etal-2023-gender, savoldi-etal-2023-test}. For instance, when translating gender-neutral person words (e.g., \emph{the students} in English) to a language with grammatical gender, the output may default to a specific gender (e.g., \emph{die Studenten (m.)} in German), thus being exclusive to other gender identities~\citep{dev-etal-2021-harms}, and 
reinforcing stereotypical biases~\citep{stanovsky-etal-2019-evaluating}.

However, the existing landscape of research on gender-fair MT is still scarce \cite{lardelli2022gender}. Previous studies are limited to covering only a few languages, scenarios, and domains---none of which focuses on German specifically. 
In this short paper, we address this gap by presenting the first study on GFL in English-to-German MT. See Figure~\ref{fig:enter-label} for an overview.

\paragraph{Contributions.} 
(\textbf{1}) We present \textsc{Gender-fair German Dictionary}, a novel resource that lists gender-neutral and gender-inclusive variants in German and their English translation. We compile this resource by enriching a community-created dictionary for German GFL. (\textbf{2}) 
We collect multi-domain data for testing the translation of gender-neutral terms from English into German in context, aligned with our dictionary.  
(\textbf{3})~We benchmark GFL in English-to-German translations involving two dedicated MT systems and six instruction-tuned models. 
We answer the following questions:

(\textbf{RQ1})~\emph{Which 
overt genders are prevalent 
in English-to-German MT outputs?}
We demonstrate that modern MT systems are 
systematically biased towards the masculine gender. GFL is extremely rare (0--2\% of all translations).

(\textbf{RQ2})~\emph{Do we observe significant differences when translating isolated words in comparison to their mentions in natural contexts?} 
Across two domains (encyclopedic and parliament speeches) we show that additional context does not yield a significantly higher portion of GFL translations.  

(\textbf{RQ3}) \emph{To what extent can the benchmarking of gender-fair German MT be automatized?} Our results show that GPT models struggle to recognize the overt gender of referents beyond the masculine and feminine forms.




\section{Background}


\subsection{Gender-Fair Language (GFL)} 

Drawing on \citet{sczesny2016can}, we use ``gender-fair'' as an umbrella term subsuming two distinct approaches: \emph{gender-neutral} and \emph{gender-inclusive} language. 
Gender-neutral describes strategies to avoid gender reference, e.g., by using passive constructions 
and gender-neutral nouns. In contrast, gender-inclusive refers to the use of different typographical characters, e.g., the interpoint ($\cdot$) in French, and symbols, e.g., schwa (\emph{\textschwa}) in Italian, 
to make all genders visible. 

\subsection{GFL Strategies for German}

In German, there are four main approaches to gender-fair language \cite{lardellitranslating}.
\emph{Gender-neutral rewording} uses passive constructions, indefinite pronouns, gender-neutral terms, or participles instead of gendered nouns.
\emph{Gender-inclusive characters} such as gender star (*), colon (:), or underscore (\_) are used to combine masculine and feminine forms as in \emph{``der*die Leser*in''} (\emph{m.}*\emph{f.} article \emph{m.}*\emph{f.} noun. Eng: 
the reader).
\emph{Gender-neutral characters and/or endings} are similar to the previous approach and include the use of \emph{``x''} or \emph{``*''} to, however, replace gender suffixes as in \emph{``dix Lesx''}.
\emph{Gender-fair neosystems} introduce a fourth gender in German alongside masculine, feminine and neuter. New pronouns, articles, and suffixes are proposed, e.g., \emph{``ens''} in \emph{``dens Lesens''}.




In this paper, we focus on strategies (1) and (2) because these are currently the most common approaches in general language use and the most likely to be adopted by professional translators~\cite{lardelli-gromann-2023-gender}.

\setlength{\tabcolsep}{15pt}
\begin{table}[!t]
\centering
\begin{small}
\begin{tabular}{@{}lll@{}}
\toprule
\textbf{Gender Form} & \textbf{Singular} & \textbf{Plural} \\ \midrule
Masculine & \emph{Berater} &\emph{Berater} \\ 
Feminine & \emph{Beraterin} &\emph{Beraterinnen} \\ 
Gender-neutral & \emph{Beratende} &\emph{Beratenden} \\ 
Gender-inclusive & \emph{Berater*in} &\emph{Berater*innen} \\ \bottomrule
\end{tabular}
\end{small}
\caption{\textbf{Dictionary entry} for \textit{``counsellor.''}}
\vspace{-0.7em}
\label{tab:seed_noun}
\end{table}

\section{Data for Gender-Fair MT}

We release two resources for studying GFL in English-to-German MT. First, we assemble a dictionary (\S\ref{ssec:dictionary}) of person-referring nouns. Second, we sample passages from Wikipedia and Europarl (\S\ref{ssec:test_corpus}) to study our terms in natural contexts.
Words in isolation allow for testing priors in translation systems, i.e., the most likely gender form for a noun when no context is provided. Natural passages enable studying the effect of contextual clues.

Note that while this work focuses on evaluating German translations, our resource can be enriched with any grammatical gender language where gender-fair language approaches ought to be preferred to masculine generics \citep[e.g.,][]{piergentili-etal-2023-gender}.

\setlength{\tabcolsep}{19pt}
\begin{table*}[!t]
\centering
\begin{small}
\begin{tabular}{@{}p{3cm}@{}p{12cm}}
\toprule
\textbf{Preceding Context} & First release On July 15, 2019, EndeavourOS released their first ISO. The team did not expect that much of the Antergos community would follow them, but the response and the numbers of community members that joined exceeded their expectations. \\ \midrule
\textbf{Matching Sentence} & Not only did the community receive the first release very well, but several \textbf{bloggers} and vloggers gave it very positive reviews, even shortly after launch. \\ \midrule
\textbf{Trailing Context} & Immediately after the launch of the distribution, the EndeavourOS team began to develop a net-installer to install with different Desktop Environments directly over the internet. \\ \bottomrule
\end{tabular}
\end{small}
\caption{\textbf{Multi-Sentence Passage from Wikipedia.} Seed noun \textit{``bloggers''} in bold. The gender of the seed is ambiguous and cannot be resolved from either context.}
\label{tab:passage_example}
\end{table*}

\subsection{Gender-Fair Dictionary} \label{ssec:dictionary}

Acknowledging the importance of hearing the voices of affected individuals in GFL research~\citep{gromann-etal-2023-participatory}, we start from the \emph{``Genderwörterbuch}.''
\footnote{\url{https://geschicktgendern.de}} This website hosts a community-created German vocabulary: users add gender-fair, usually neutral, alternatives to commonly gendered terms. Next, we sample and select suitable terms for our research, and further enrich the dictionary.

\paragraph{Term Selection.} We start from 128 randomly selected terms. We filter out those that were already neutral, e.g., \emph{``Star,''} which is an Anglicism and does not have variants for other genders in German. To facilitate back-translation into English, we remove polysemous terms, e.g., \emph{``aid.''}

\paragraph{Dictionary Enrichment.} One of the authors---experienced with GFL and translation---enriched every noun with its masculine, feminine, gender-inclusive, and gender-neutral form in singular and plural. 
We use gender star (*) for gender-inclusive forms, as it is common in German-speaking countries \cite{korner2022gender}.
Finally, we manually translated each term into English.
Our final dictionary counts 115 nouns in their singular and plural forms (see Table~\ref{tab:seed_noun} for an example). 
Notably, the final list contains both professions (e.g., \emph{``deputy''}) as well as common nouns (e.g., \emph{``donor''}). While, to date, most research on gender bias in MT focused on the translation of profession terms only 
\citep{prates2020assessing},
we expand the focus and include common nouns referring to people in a broader sense. 



\subsection{Multi-Sentence Multi-Domain Mentions} \label{ssec:test_corpus}

We collect an additional set of English passages that mention our dictionary entries in their plural form. We focus on plural occurrences because they yield to gender-ambiguous cases more frequently, providing a more challenging scenario for translation systems.

 
\paragraph{Data Sources.} We sample sentences from Europarl \citep{koehn-2005-europarl} and Wikipedia.\footnote{We use Europarl's release v7 (parallel corpus English-German) and the Wikipedia snapshot at 01--03--2022 at \url{https://huggingface.co/datasets/wikipedia}.} 
Europarl is a widely recognized benchmark dataset for MT displaying institutional language from parliamentary speeches---perhaps amongst the first contexts GFL was devised for \cite{piergentili-etal-2023-hi}.
Wikipedia presents encyclopedic text, opening to new contexts where our seed nouns appear.  

\paragraph{Passage Retrieval.} For each of our 115 terms, we retrieve all sentences in a given corpus with at least one occurrence of the noun.\footnote{We use \texttt{nltk} to split paragraphs into sentences. Since several words can be used as adjectives, we extract POS tags with \texttt{spacy}'s morphological utility and match only NOUNs.}
The seed's gender assignment might require cross-sentence resolution. Thus, limited to Wikipedia, we extract the matching sentence along with two preceding sentences and one following (see Table~\ref{tab:passage_example}).

Concretely, we sample passages in two steps. First, we randomly selected five passages per seed noun, yielding an initial batch of 358 single-sentence passages from Europarl and 400 multi-sentence passages from Wikipedia. Respectively, 36 and 35 seeds did not match any sentence in Europarl and Wikipedia, and we matched only one or two sentences for some seeds.
Then, we manually filtered passages via quality checks on the matching sentence. Specifically, we ensure that (i) the overt gender of the seed words is ambiguous or it refers to a mixed-gender group, (ii) the passage meaning is self-contained, and (iii) the passages do not exceed a length of 100 words.\footnote{The average passage length is 34 and 92 words for Europarl and Wikipedia, respectively.} 

\section{Experiments}
\label{sec:experiments}

\subsection{Translation System Selection}
Acknowledging that, today, people are exposed to MT in multiple ways, we include in our study a variety of systems. As commercial representatives of dedicated MT systems, we include Google Translate and DeepL. Additionally, we study GPT 3.5 and GPT 4 \citep{openai2023gpt4}, accessible through online APIs. We also include open-weight models: two supervised MT models, NLLB \citep{costa2022no} and OPUS MT \citep{tiedemann-thottingal-2020-opus}, Flan-T5 \citep{chung2022scaling}, a multi-task instruction fine-tuned model, and Llama 2 \citep{touvron2023llama2}. See Appendix~\ref{sec:details_translation_systems} for full details. 

\subsection{Translation and Evaluation} \label{ssec:translation_evaluation}
We machine-translated all seed words in isolation (singular and plural) and the passages retrieved from Europarl and Wikipedia (\S\ref{ssec:test_corpus}). 
The same author from \S\ref{ssec:dictionary} manually annotated whether the term is translated with a masculine, feminine, gender-inclusive, or gender-neutral form. 
Since words in isolation were sometimes mistranslated, we noted the type of errors. 
Mistakes were due to semantics (i.e., the German term has a different meaning than the English source), grammar (e.g., wrong number or no agreement between article and nouns), and hallucinations.
We finally annotate three out of the five passages retrieved from Europarl and Wikipedia with the same criteria.\footnote{The two passages excluded contain complex phrasing, formatting problems---e.g., Wikipedia section titles, which are not proper preceding context---or severe translation mistakes.} The final number of annotated passages is 215 and 218 for Europarl and Wikipedia, respectively. In order to validate our analysis, a German native speaker student research assistant with previous experience in MT output annotation repeated the analysis on a the OPUS MT's outputs in the singular and plural, and the GPT3.5's translations of the Wikipedia passages. 

\setlength{\tabcolsep}{17pt}
\begin{table}[!t]
\centering
\begin{small}
\begin{tabular}{@{}llll@{}}
\toprule
 & \textbf{\# SN} & \textbf{\# RP} & \textbf{\# AP} \\ \midrule
\textbf{Europarl} & 79 & 358 & 215 \\
\textbf{Wikipedia} & 80 & 400 & 218 \\ \bottomrule
\end{tabular}
\end{small}
\caption{\textbf{\textsc{Gender-fair German Dictionary} statistics.} Number of seed nouns (SN), and retrieved (RP) and annotated passages (AP) from Europarl and Wikipedia.}
\vspace{-0.7em}
\label{tab:corpus_statistics}
\end{table}

Additionally, to answer \textbf{RQ3}, we prompt GPT 3.5 in zero-shot to detect GFL in the translations (see Appendix~\ref{sec:automatic_eval} for details). We compare these results with the manual annotations.





\subsection{Results}
\noindent\textbf{Words-in-Isolation.}
As shown in Table \ref{tab:words_in_isolation_singular} (see Appendix~\ref{sec:detailed_results}), \textbf{all models are heavily biased towards masculine forms} (93--96\% of all translations, (\textbf{RQ1})). MT systems use feminine forms seldom (2--4\%), usually when nouns relate to professions that are stereotypically associated with femaleness like \emph{``children's day carer,''} \emph{``kindergarten teacher,''} and \emph{``secretary''}. Gender-neutral and inclusive forms are even rarer (0--2\%). One example is the term \emph{``newcomer,''} translated by nearly all models with the gender-neutral \emph{``Neuling''} that is very common in German. While its grammatical gender is masculine, it is used for all genders. Interestingly, Flan-T5 produced many mistranslations. For instance, the seed noun \emph{``traveller''} was translated to \emph{``Reisenden''} with a grammatical mistake in the noun declension, i.e., the suffix ``n.'' The model also created non-existing words, e.g., \emph{``Antwortent''} for \emph{``respondent,''} instead of \emph{``Befragten''} (``Antwort'' is ``answer'' in English).

\begin{figure}[!t]
    \centering
    \includegraphics[width=0.99\linewidth]{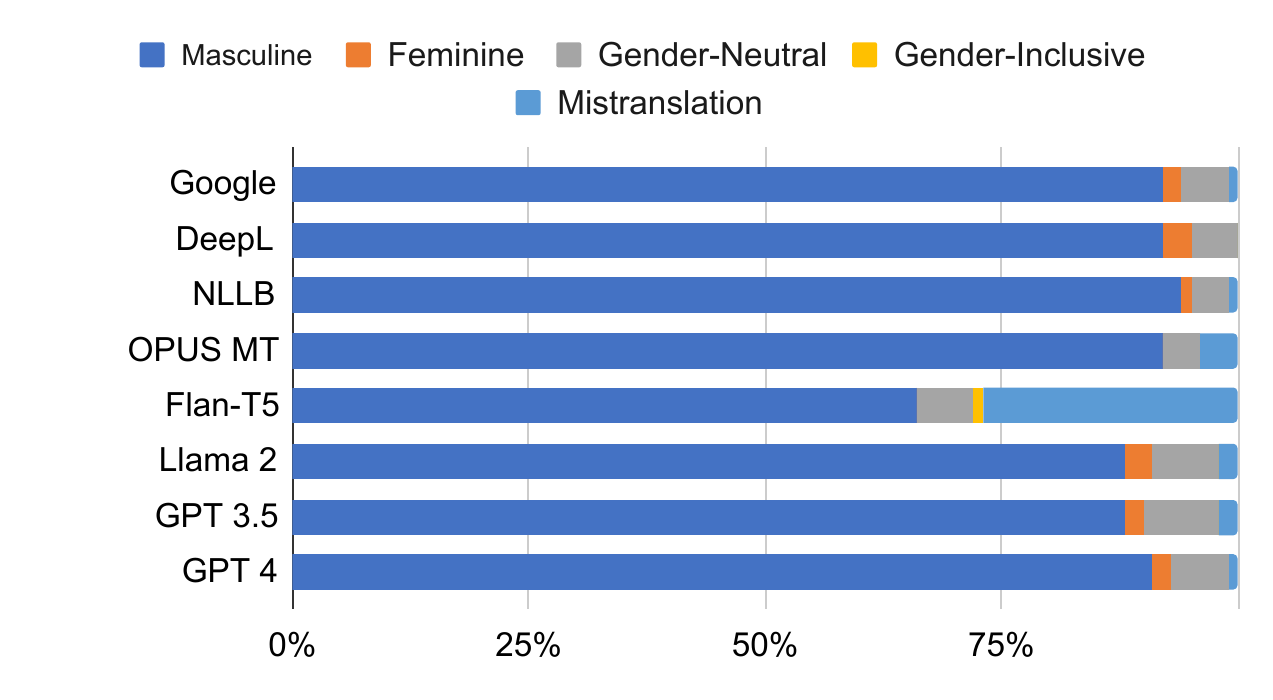}
    \caption{\textbf{Results for plural words in isolation.} Gender form distribution and mistranslations for each translation system.}
    \label{fig:words_isolation_plural}
    \vspace{-0.5em}
\end{figure}

The analysis of plural translations yielded similar results (see Figure \ref{fig:words_isolation_plural}). Gender-neutral forms occur slightly more frequently (4--8\% of all translations), probably because of two reasons.
First, while some nouns, e.g., \emph{``practitioner''} are gender-specific in the singular (\emph{``Praktiker''/``Praktikerin''}), gender-neutral alternatives are common for plural (\emph{``Praktizierende''}). Second, some nouns have the same form for masculine and feminine but the article is gender-specific in the singular only, e.g., \emph{``the relative''} (\emph{``die Angehörige''}/ \emph{``die Angehörigen''}). We report full results in Table~\ref{tab:words_in_isolation_plural}, Appendix~\ref{sec:detailed_results}.\footnote{A second annotator (see \ref{ssec:translation_evaluation}) annotated OPUS MT's outputs in the singular and plural. Agreement on gender forms was perfect, with a Cohen's kappa of 1.00.}
 


\paragraph{Words-in-Context.} For answering \textbf{RQ2}, we conducted a focused analysis on GPT 3.5 and DeepL translations of the passages retrieved from Europarl and Wikipedia.
These models produced the highest number of non-masculine translations among MT systems and language models.
\begin{figure}[t]
    \centering
    \includegraphics[width=0.95\linewidth]{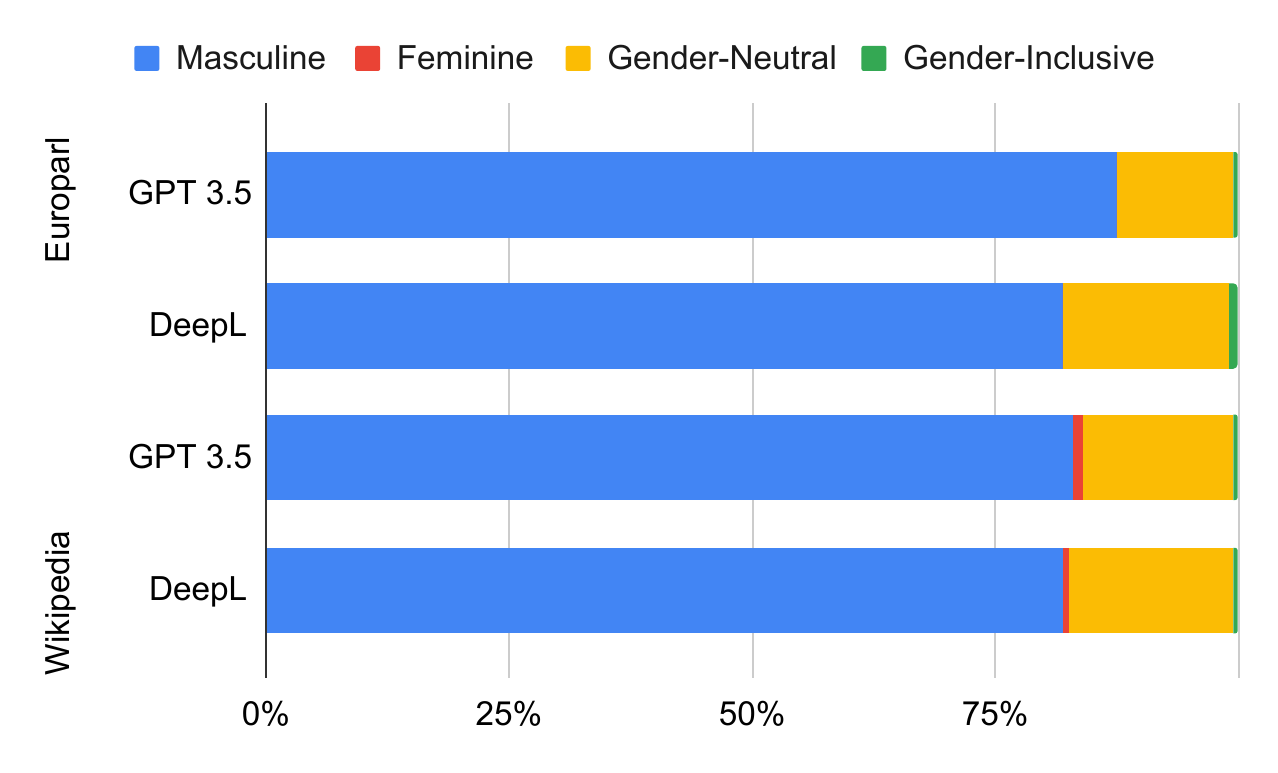}
    \caption{\textbf{Results for words in context (plural).} Gender form distribution in GPT 3.5 and DeepL translations for each data source.}
    \label{fig:words_context}
    \vspace{-0.5em}
\end{figure}
The results are shown in Figure \ref{fig:words_context} and absolute frequencies are reported in Table \ref{tab:words_context} in Appendix~\ref{sec:detailed_results}. \textbf{Both models are strongly biased towards masculine forms} (85\% of all translations). While feminine and gender-inclusive forms are rare (about 1\% of cases), gender-neutral forms are more common (\textasciitilde 15\%). Systems use them for nouns that are already gender-neutral (e.g., \emph{``travellers''}, \emph{``respondents''}, and \emph{``relatives''}), or for which a gender-neutral alternative is common in the plural (e.g., \emph{``practitioners''}, \emph{``chairpeople''}, \emph{``newcomers}'').\footnote{In this case, the second rater repeated the analysis on a portion of data, i.e., GPT 3.5's translations of the Wikipedia passages. Cohen's kappa was 0.954. The few disagreements were mistakes that were corrected.}


\paragraph{Zero-shot GFL Detection.} We test whether GPT 3.5 and GPT 4 can serve as viable tools for automatic detection of GFL. To this end, we prompted the models to label the translations of words in context produced with GPT 3.5 and compare the results with our manual annotations. 
Note that we found no feminine forms in this set of translation.

Table~\ref{tab:gpt_eval_performance} reports agreement results with human evaluation.
Both GPT 3.5 and GPT 4 achieve an extremely low recall (11.5\%) for gender-neutral cases. However, GPT 4's precision is relatively high (75\%) compared to GPT 3.5 (30\%), showing an improvement model generations.  
These findings highlight that \textbf{zero-shot automatic detection of GFL in German with recent GPT models is hard}, and underscore the importance of expert human oversight when studying GFL in MT.  


\setlength{\tabcolsep}{14pt}
\begin{table}[!t]\centering
\small
\begin{tabular}{@{}lrrrr}\toprule
\textbf{Gender} &\textbf{P} &\textbf{R} &\textbf{S} \\\midrule
Masculine &92.9 &69.7 &188 \\
Feminine & - & - &0 \\
Gender-Inclusive &4.8 &100 &1 \\
Gender-Neutral &30.0 &11.5 &26 \\ \midrule \midrule
Masculine &96.3 &96.3 &188 \\
Feminine & - & - & 0 \\
Gender-Inclusive &6.2 &100 &1 \\
Gender-Neutral &75.0 &11.5 &26 \\
\bottomrule
\end{tabular}
\caption{\textbf{Automatic detection of GFL.} (P)recision, (R)ecall, and (S)upport of GPT 3.5 (top) and GPT 4 (bottom) zero-shot labeling when compared to human analysis. Europarl \textsc{En-De} (n=215).}\label{tab:gpt_eval_performance}
\vspace{-1em}
\end{table}

\section{Related Work}

Due to stereotypical and exclusive biases present in the training data, the output of MT may discriminate against certain genders \cite[e.g.,][]{stanovsky-etal-2019-evaluating, attanasio-etal-2023-tale}. In this context, recent research has focused on the issue of gender exclusivity~\citep{piergentili-etal-2023-gender}. Towards a better understanding, much attention has been paid to studying existing strategies chosen by human subjects, like translation team leaders~\cite{daems-2023-gender}, and MT post editors~\cite{lardelli-gromann-2023-gender, paolucci-etal-2023-gender}. Related to this, \citet{gromann-etal-2023-participatory} pointed to participatory research as a promising avenue. Another research thread focuses on assessing the capabilities of existing MT systems:  \citet{lauscher-etal-2023-em} investigated the translation of pronouns in commercial MT, 
\citet{saunders-olsen-2023-gender} the translation of named entities, and 
\citet{piergentili-etal-2023-hi} benchmarked gender-neutral MT from English to Italian. \citet{savoldi-etal-2023-test} report the results of a shared task, designed to assess the GFL ability of MT systems from German to English.
The only existing work, which also focuses, like ours, on English to German GFL is \citet{kostikova-etal-2023-adaptive}.
However, the authors study 15 sentences only. In contrast, we focus on 115 words in multiple translation scenarios.



\section{Conclusion}
We have presented the first study on gender-fair \textsc{En}-\textsc{De} MT. We introduced two novel resources grounded in community contributions. We experimented with eight translation systems and several setups, including words in isolation, natural passages from the encyclopedic domain and parliamentary speeches. Our findings call for more research on GFL in modern MT towards fairer and more inclusive translation technology.

\section*{Acknowledgments}

Giuseppe Attanasio was supported by the Portuguese Recovery and Resilience Plan through project C645008882-00000055 (Center for Responsible AI) and by Fundação para a Ciência e Tecnologia through contract UIDB/50008/2020. He conducted part of the work as a member of the MilaNLP group at Bocconi University, Milan.

\section*{Limitations}

This work comes with several limitations.

We focus on a single language pair and direction, English->German. Our choice is dictated by the lack of consensus for gender-fair language \cite{ackerman2019syntactic}. German is a notable exception where seminal work has been recently conducted. Hence, we opted to limit the scope and study whether MT systems keep up with the growing trends.

Our study is limited to a relatively small number of seed nouns and sampled sentences. We acknowledge this aspect but highlight that our procedure generalizes easily to new seeds and data sources. Moreover, when sampling natural passages, we did not control for specific factors related to gender and gender inflection. First, we focus on sentences where the entity's gender is ambiguous or mixed. Therefore, we discard all cases where the entity's gender is disambiguated, for example, by lexical clue within the matching sentence. Second, we do not control for the presence of other human entities that might act as a confounding factor.  



\section*{Ethical Considerations}
By investigating gender in MT, our work focuses on the exclusionary potential of language technologies which might impact the visibility and/or mental health of minoritized groups such as women and non-binary people \cite {sczesny2016can,mclemore2018minority}. Here, we also enrich a community-created GFL dictionary. Since there is no acknowledged standard for GFL, the alternatives we present in our work are not prescriptive, though they represent common strategies.

\bibliography{anthology,custom}

\appendix

\section{Details on Translation Systems}
\label{sec:details_translation_systems}

We used paid APIs and \texttt{deep-translator}\footnote{\url{https://github.com/nidhaloff/deep-translator}} for Google Translate and DeepL, and accessed \texttt{gpt-3.5-turbo-0613} (GPT 3.5) and \texttt{gpt-4-0613} (GPT 4). For all open-weight models, we used code and implementation from \texttt{transformers} \citep{wolf-etal-2020-transformers} and \texttt{simple-generation} \citep{milanlp-2023-simple-generation} as the inference engine. In particular, we used \href{https://huggingface.co/Helsinki-NLP/opus-mt-en-de}{Helsinki-NLP/opus-mt-en-de} (OPUS MT), \href{https://huggingface.co/facebook/nllb-200-3.3B}{facebook/nllb-200-3.3B} (NLLB), \href{https://huggingface.co/google/flan-t5-xxl}{google/flan-t5-xxl} (Flan-T5), and \href{https://huggingface.co/meta-llama/Llama-2-70b-chat-hf}{meta-llama/Llama-2-70b-chat-hf} (Llama 2).

To run the experiments, we used an in-house computing center and run all the experiments on one A100 GPU. 

\paragraph{Prompt and Decoding.} We used no prompts from supervised MT models, whereas for Llama 2 and GPTs we used:\\

\noindent
\texttt{Translate the following sentence into German. Reply only with the translation. Sentence: \{sentence\}} \\

Finally, we followed FLAN's \citep{longpre2023flan} translation templates for Flan-T5: \\

\noindent
\texttt{\{sentence\}$\backslash$n$\backslash$nTranslate this into German?} \\

We used the default generation configuration for GPTs, beam search decoding (n=5) for OPUS MT, NLLB, and Flan-T5, and nucleus sampling (top p=1, top k=50, temperature=0) for Llama 2. 

\section{Automatic Evaluation}
\label{sec:automatic_eval}

We prompted GPT 3.5 and GPT 4 with default decoding parameters to evaluate whether machine translated passages used any gender-fair form.

The prompt we used is:\\

\noindent
\texttt{If the following sentence contains the German translation for the English word \{seed\_noun\}, tell me which overt gender it displays among Masculine, Feminine, Gender-Neutral, or Gender-Inclusive. If no translation is found, reply with None. Sentence: \{translation\}} \\

We substituted \texttt{seed\_noun} and \texttt{translation} accordingly.

\section{Detailed Results}
\label{sec:detailed_results}

Tables~\ref{tab:words_in_isolation_singular}--\ref{tab:words_context} report the total occurrences of different gender forms and mistranslations in our setups.

\begin{table*}[!t]
\centering
\small
\begin{tabular}{p{2cm}p{1cm}p{1cm}p{1cm}p{1cm}p{1cm}p{1cm}}
\toprule
\multirow{2}{*}{\textbf{Model}} & \multicolumn{4}{l}{\textbf{Gender}}                                                          & \multirow{2}{*}{\textbf{Mi}} \\ \cmidrule(lr){2-5}
                                & \textbf{M} & \textbf{F} & \textbf{GN} & \textbf{GI} &                                          \\ \toprule
DeepL   & 108 & 5 & 1  & 1 & 0  \\
GT    & 107 & 3 & 1  & 0 & 4  \\
GPT 3.5 & 107 & 2 & 1 & 0 & 5  \\
GPT 4   & 108 & 2 & 1  & 0 & 4  \\
NLLB    & 111 & 2 & 1  & 0 & 1  \\
OPUS MT & 110 & 3 & 0  & 0 & 0  \\
Flan-T5 & 51  & 9 & 3  & 0 & 39 \\
Llama 2 & 107 & 3 & 2  & 0 & 2  \\ \bottomrule
\end{tabular}
\caption{\textbf{Results of the words in isolation analysis (singular).} For each seed word, we count masculine (M), feminine (F), gender-neutral (GN), and gender-inclusive forms (GI), and mistranslations (Mi).}
\label{tab:words_in_isolation_singular}
\end{table*}

\begin{table*}[!t]
\centering
\small
\begin{tabular}{p{2cm}p{1cm}p{1cm}p{1cm}p{1cm}p{1cm}p{1cm}}
\toprule
\multirow{2}{*}{\textbf{Model}} & \multicolumn{4}{l}{\textbf{Gender}}                                                          & \multirow{2}{*}{\textbf{Mi}} \\ \cmidrule(lr){2-5}
                                & \textbf{M} & \textbf{F} & \textbf{GN} & \textbf{GI} &                                          \\ \toprule
DeepL   & 106 & 3 & 6  & 0 & 0  \\
GT    & 105 & 2 & 6  & 0 & 1  \\
GPT 3.5 & 101 & 2 & 10 & 0 & 2  \\
GPT 4   & 105 & 2 & 7  & 0 & 1  \\
NLLB    & 108 & 1 & 5  & 0 & 1  \\
OPUS MT & 105 & 0 & 5  & 0 & 5  \\
Flan-T5 & 76  & 0 & 7  & 1 & 31 \\
Llama 2 & 101 & 3 & 8  & 0 & 2  \\ \bottomrule
\end{tabular}
\caption{\textbf{Results of the words in isolation analysis (plural).} For each seed word, we count masculine (M), feminine (F), gender-neutral (GN), and gender-inclusive (GI) forms, and mistranslations (Mi).}
\label{tab:words_in_isolation_plural}

\end{table*}

\begin{table*}[!t]
\centering
\small
\begin{tabular}{@{}llllll@{}}
\toprule
\multirow{2}{*}{\textbf{Source}} &
  \multirow{2}{*}{\textbf{Model}} &
  \multicolumn{4}{l}{\textbf{Gender}} \\ \cmidrule(l){3-6} 
 &
   &
  \multicolumn{1}{l}{\textbf{M}} &
  \multicolumn{1}{l}{\textbf{F}} &
  \multicolumn{1}{l}{\textbf{GN}} &
  \multicolumn{1}{l}{\textbf{GI}} \\ \midrule
Europarl  & GPT 3.5 & 188 & 0 & 26 & 1 \\
                   & DeepL   & 177 & 0 & 37 & 1 \\ \midrule
Wikipedia & GPT 3.5 & 181 & 2 & 34 & 1 \\
                   & DeepL   & 178 & 1 & 38 & 1 \\ \bottomrule
\end{tabular}
\caption{\textbf{Results of the words in context analysis (plural).} For each seed word, we count masculine (M), feminine (F), gender-neutral (GN), and gender-inclusive (GI) forms.}
\label{tab:words_context}

\end{table*}

\end{document}